\documentclass[11pt,a4paper]{article}
\usepackage[hyperref]{emnlp2018}
\usepackage{times}
\usepackage{latexsym}

\usepackage{url}
\usepackage{times}
\usepackage{xcolor}
\usepackage[utf8]{inputenc}
\usepackage[small]{caption}
\usepackage{graphicx}
\usepackage{booktabs}
\usepackage{multirow}
\usepackage{mathrsfs}
\usepackage{amsmath}
\usepackage{hyperref}
\usepackage{amssymb}
\usepackage{balance}
\usepackage{graphicx}
\usepackage{caption}
\usepackage{subcaption}
\usepackage{balance}

\newcommand{\reffig}[1]{Figure \ref{#1}}
\newcommand{\reftbl}[1]{Table \ref{#1}}
\newcommand{\refsec}[1]{Section \ref{#1}}

\newcommand{\m}[1]{\mathcal{#1}}
\newcommand{\AND}{Attentive NeuralDater}
\newcommand{\method}{AD3}
\newcommand{\methodsyntac}{Attentive Context Model}
\newcommand{\methodtemp}{Ordered Event Model}
\newcommand{\methodsyntacshort}{AC-GCN}
\newcommand{\methodtempshort}{OE-GCN}

\newcommand*{\Scale}[2][4]{\scalebox{#1}{$#2$}}%
\aclfinalcopy % Uncomment this line for the final submission
\title{\method{}: Attentive Deep Document Dater}

\author{
	Swayambhu Nath Ray \\
	Indian Institute of Science \\
	Bangalore, India\\
	{\small {\tt swayambhunath93@gmail.com}} \And
	Shib Sankar Dasgupta \\
	Indian Institute of Science \\
	Bangalore, India\\
	{\small {\tt s.s.dasgupta.iisc@gmail.com}} \And
	Partha Talukdar \\
	Indian Institute of Science \\
	Bangalore, India\\
	{\small {\tt ppt@iisc.ac.in}}
}
\date{}
\begin{document}
	\maketitle
	\begin{abstract}
Knowledge of the creation date of documents facilitates several tasks such as summarization, event extraction, temporally focused information extraction etc. Unfortunately, for most of the documents on the Web, the time-stamp metadata is either missing or can't be trusted. Thus, predicting creation time from document content itself is an important task. %but relatively unexplored problem. %Apart from previous statistical and feature engineered methods, there exists a deep learning based approach for this task. 
In this paper, we propose Attentive Deep Document Dater (\method{}), an attention-based neural document dating system which utilizes both context and temporal information in documents in a flexible and principled manner. We perform extensive experimentation on multiple real-world datasets to demonstrate the effectiveness of \method{} over neural and non-neural baselines.
	\end{abstract}
	
	% !TeX spellcheck = en_US
\section{Introduction}
\label{sec:introduction}

Many natural language processing tasks require document creation time (DCT) information as a useful additional metadata. Tasks such as  information retrieval \cite{ir_time_li,ir_time_dakka}, temporal scoping of events and facts  \cite{event_detection,talukdar2012coupled}, document summarization \cite{text_summ_time} and analysis \cite{history_time} require precise and validated creation time of the documents. Most of the documents obtained from the Web either contain DCT that cannot be trusted or contain no DCT information at all \cite{Kanhabua:2008:ITL:1429852.1429902}. Thus, predicting the time of these documents based on their content %, by using a few annotated documents 
is an important task, often referred to as \emph{Document Dating}.

%Although a few generative modeling approaches \cite{de_jong05, Kanhabua:2008:ITL:1429852.1429902}  exists for this task, discriminative model proposed by \citet{Chambers:2012:LDT:2390524.2390539} outperforms all the previous models. \citet{Kotsakos:2014:BAD:2600428.2609495} deploys term-burstiness and gains a significant boost in precision.

A few generative approaches \cite{de_jong05, Kanhabua:2008:ITL:1429852.1429902} as well as a discriminative model \cite{Chambers:2012:LDT:2390524.2390539} have been previously proposed for this task. \citet{Kotsakos:2014:BAD:2600428.2609495} employs term-burstiness resulting in improved precision on this task.

Recently proposed NeuralDater \cite{NeuralDater} uses a graph convolution network (GCN) based approach for document dating, outperforming all previous models by a significant margin. NeuralDater extensively uses the syntactic and temporal graph structure present within the document itself. Motivated by NeuralDater, we explicitly develop two different methods: a) \emph{\methodsyntac{}}, and b) \emph{\methodtemp{}}. The first component tries to accumulate knowledge across documents, whereas the latter uses the temporal structure of the document for predicting its DCT.

Motivated by the effectiveness of attention based models in different NLP tasks \cite{attention_qna,bahdanau_attn}, we incorporate attention in our method in a principled fashion. We use attention not only to capture context but also for feature aggregation in the graph convolution network \cite{graphsage}. %We not only did some focused context capturing using attention but also incorporated it for feature aggregation \cite{graphsage} for the graph convolution network. 
Our contributions are as follows. 

\begin{itemize}
	\item We propose Attentive Deep Document Dater (\method{}), the first attention-based neural model for time-stamping documents.
	\item We devise a novel method for label based attentive graph convolution over directed graphs and use it for the document dating task.
	\item Through extensive experiments on multiple real-world datasets, we demonstrate \method{}'s effectiveness over previously proposed  methods.
\end{itemize}
\begin{figure*}[!t]
	\centering
	\fbox{\includegraphics[width=6in]{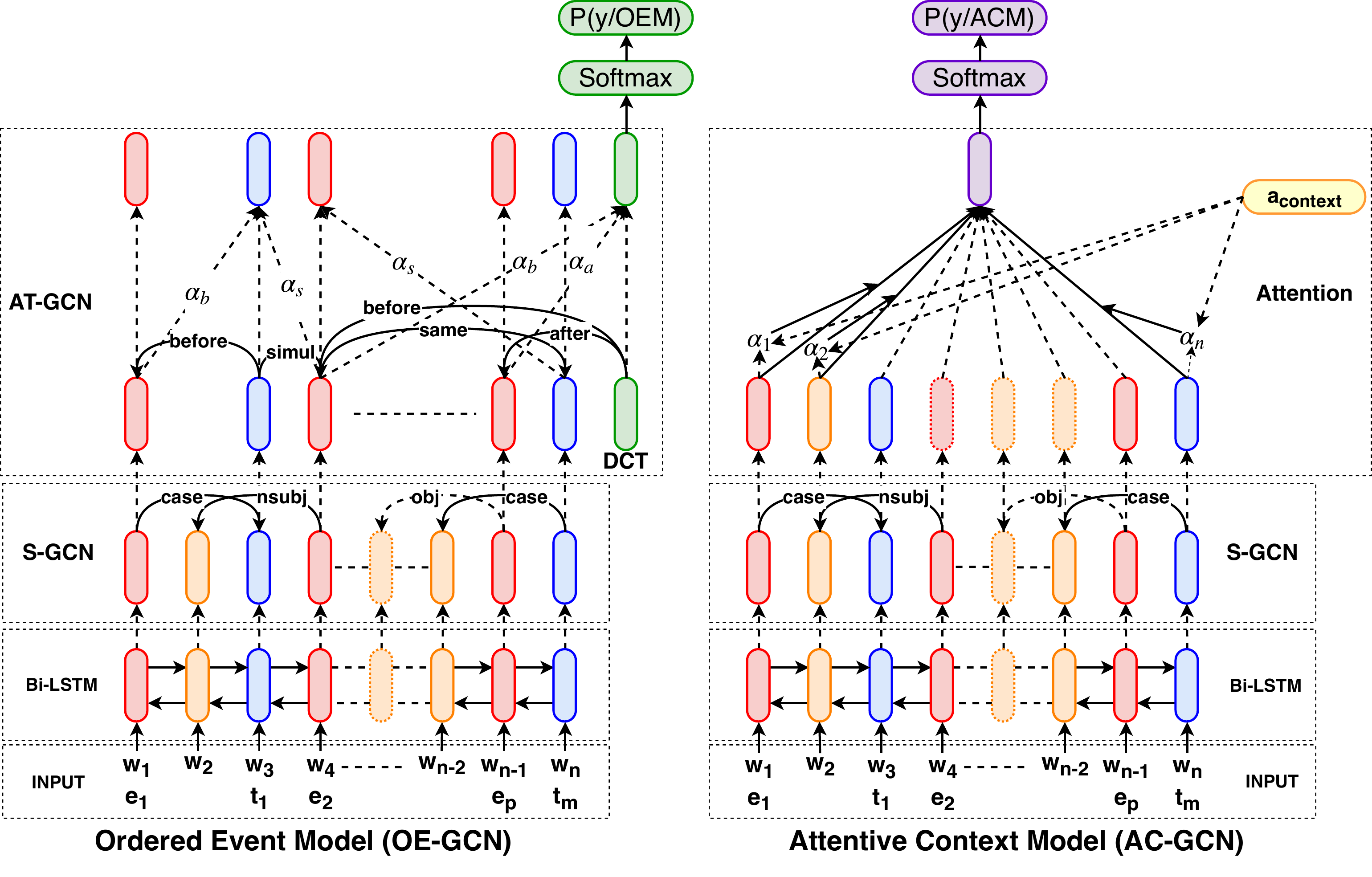}}
	\caption{\label{fig:overview} Two proposed models a) \methodtemp{} (left) and b) \methodsyntac{} (right), where $\text{w}_{\text{i}}$ are the words of a document (D), $\text{e}_{\text{i}}$ are the words signifying events and $\text{t}_{\text{i}}$ are the temporal tokens as detected in the document. Both models use Bi-LSTM and S-GCN (Syntactic-GCN, see \refsec{sec: syntactic_embedding}) in the initial part of their pipeline. \methodtemp{} (\methodtempshort{}) uses a label based attentive graph convolutional network for encoding the DCT, whereas \methodsyntac{} (\methodsyntacshort{}) uses a word attention based model to encode the document. $\alpha_{i} (\forall \ i \in [1,n])$ denotes attention  over the words of document and $\alpha_{a}, \ \alpha_{b} $ and $\alpha_{s}$ denote attention over nodes connected with edge labels \textit{AFTER}, \textit{BEFORE} and \textit{SIMULTANEOUS},  respectively. \methodtempshort{} provides the probability scores over the years given the encoded DCT, while \methodsyntacshort{} provides the probability scores given the context of the document. Both the models are trained separately.}
	
\end{figure*}

%\begin{itemize}
%    \item Attempts made in previous works for this problem.
%    \item Some introduction about the CATENA paper.
%    \item Some introduction about graph convolution networks and attentive graph convolution networks.
%    \item Some example with graph attention and how attention is helping.
%    \item Main Attentive Neural Dater diagram on the 2nd page.
%\end{itemize}5

AD3 source code and datasets used in the paper are available at \href{https://github.com/malllabiisc/AD3}{https://github.com/malllabiisc/AD3}
	% !TeX spellcheck = en_GB
\section{Related Work}
\label{sec:related_work}
	{\bf Document Time-Stamping:}
	Initial attempts made for document time-stamping task include statistical language models proposed by \citet{de_jong05} and \citet{Kanhabua:2008:ITL:1429852.1429902}. \cite{Chambers:2012:LDT:2390524.2390539} use temporal and hand-crafted features extracted from documents to predict DCT. They propose two models, one of which learns the probabilistic constraints between year mentions and the actual creation time, whereas the other one is a discriminative model trained on hand-crafted features. \citet{Kotsakos:2014:BAD:2600428.2609495} propose a term-burstiness \cite{Lappas:2009:BSD:1557019.1557075} based statistical method for the task. \citet{NeuralDater} propose a deep learning based model which exploits the temporal and syntactic structure in documents using graph convolutional networks (GCN).
	
	{\bf Event Ordering System:}
	The task of extracting temporally rich events and time expressions and ordering between them is introduced in the TempEval challenge \cite{tempeval13,tempeval10}. Various approaches \cite{caveo_plus, catena_paper} made for solving the task use sieve-based architectures, where multiple classifiers are ranked according to their precision and their predictions are weighted accordingly resulting in a temporal graph structure. A method to extract temporal ordering among relational facts was proposed in \cite{talukdar2012acquiring}.
	
	{\bf Graph Convolutional Network (GCN):}
	GCN \cite{kipf2016semi} is the extension of convolutional networks over graphs. In different NLP tasks such as semantic-role labeling \cite{gcn_srl}, neural machine translation \cite{gcn_nmt}, and event detection \cite{gcn_event}, GCNs have proved to be effective. We extensively use GCN for capturing both syntactic and temporal aspect of the document.
	
	{\bf Attention Network:}
	Attention networks have been well exploited for various tasks such as  document classification \cite{hierarchical_attn}, question answering \cite{attention_qna}, machine translation \cite{bahdanau_attn, NIPS2017_7181}. Recently, attention over graph structure has been shown to work well by \citet{graphAttention}. Taking motivation from them, we deploy an attentive convolutional network on temporal graph for the document dating problem.
	% !TeX spellcheck = en_GB

\section{Background: GCN \& NeuralDater}
\label{sec:background}
The task of document dating can be modeled as a multi-class classification problem. Following prior work, we shall focus on DCT prediction at the year-granularity in this paper. % where the previous work tries to predict the DCT at a year-wise granularity. 
In this section, we summarize the previous state-of-the-art model NeuralDater \cite{NeuralDater}, before moving onto our method. An overview of graph convolutional network (GCN) \cite{kipf2016semi} is also necessary as it is used in NeuralDater as well as in our model.

\subsection{Graph Convolutional Network}
\label{sec:GCN}
\textbf{GCN for Undirected Graph:} 
Consider an undirected graph, $\m{G} = (\m{V}, \m{E})$, where $\m{V}$ and $\m{E}$ are the set of  $n$ vertices and set of edges respectively. Matrix  $\m{X} \in \mathbb{R}^{n \times m}$, whose rows are input representation of node $u$, where $x_{u} \in \mathbb{R}^{m}\text{, }\ \forall \ u \in \m{V}$,  is the input feature matrix. The output hidden representation $h_v \in \mathbb{R}^{d}$ of a node $v$ after a single layer of graph convolution operation can be obtained by considering only the immediate neighbours of  $v$, as formulated in \cite{kipf2016semi}. In order to capture information at multi-hop distance, one can stack layers of GCN, one over another. 
%In particular,  $h^{k+1}_{v}$, representation of node $v$  after $k^{th}$ GCN layer can be formulated as:
%%
%$$h^{k+1}_{v} = f\left(\sum_{u \in \m{N}(v)}\left(W^{k} h^{k}_{u} + b^{k}\right)\right), \forall v \in \m{V} . $$
%%
%Output feature matrix $\m{H} \in \mathbb{R}^{k \times n}$ composed of $h_{v} \in \mathbb{R}^{k}\text{, }\forall v \in \m{V}$, encapsulates the immediate neighboring node features i.e. hidden representation for each $v$ encompasses the input feature of the set \{$u : \text{  } u \in \m{N}(v) $\}.

\noindent \textbf{GCN for Directed Graph:}
\label{sec:directed_gcn}
Consider a labelled edge from node $u$ to $v$ with label $l(u, v)$, denoted collectively as $(u, v, l(u, v))$. Based on the assumption that information in a directed edge need not only propagate along its direction, \citet{gcn_srl} added opposite edges viz., for each $(u, v, l(u, v))$,  $(v,u,l(u,v)^{-1})$ is added to the edge list. Self loops are also added for passing the current embedding information.
%\begin{multline}    
%\m{E'} = \m{E} \cup \{(v,u,l(u,v)^{-1})~|~ (u,v,l(u,v)) \in \m{E}\}  \\
%\cup \{(u, u, \top)~|~u \in \m{V})\} 
%\label{eqn:updated_edges}.
%\end{multline}
When GCN is applied over this modified directed graph, the embedding of the node $v$ after $k^{th}$ layer will be, 
\begin{equation*}
h_{v}^{k+1} = f \left(\sum_{u \in \m{N}(v)}\left(W^{k}_{l(u,v)}h_{u}^{k} + b^{k}_{l(u,v)}\right)\right).
\label{eqn:gcn_diretected}
\end{equation*}

We note that the parameters $W^{k}_{l(u,v)}$ and $b^{k}_{l(u,v)}$ in this case are edge label specific. $h^{k}_{u}$ is the input to the $k^{th}$ layer. Here, $\m{N}(v)$ refers to the set of neighbours of $v$, according to the updated edge list and $f$ is any non-linear activation function (e.g., ReLU: $f(x) = \max(0, x)$).

\subsection{NeuralDater}
\label{sec:neural_dater}
In this sub-section, we provide a brief overview of the components of the NeuralDater \cite{NeuralDater}. 
Given a document $D$ with $n$ tokens $w_{1}, w_{2}, \cdots w_{n}$, NeuralDater extracts a temporally rich embedding of the document in a principled way as explained below:
\subsubsection{Context Embedding}
\label{sec:context_lstm}
Bi-directional LSTM is employed for embedding each word with its context. The GloVe representation of the words $X \in \mathbb{R}^{n\times k}$ is transformed to a context aware representation $H^{cntx} \in \mathbb{R}^{n \times k}$ to get the context embedding. This is essentially shown as the Bi-LSTM in \reffig{fig:overview}.

\subsubsection{Syntactic Embedding}
\label{sec: syntactic_embedding}

In this step, the context embeddings are further processed using GCN over the dependency parse tree of the sentences in the document, in order to capture long range connection among words. The syntactic dependency structure is extracted by Stanford CoreNLP’s dependency parser \cite{stanford_corenlp}. NeuralDater follows the same formulation of GCN for directed graph as described in \refsec{sec:directed_gcn}, where additional edges are added to the graph to model the information flow. Again following \cite{gcn_srl}, NeuralDater does not allocate separate weight matrices for different types of dependency edge labels, rather it considers only three type of edges: a) edges that exist originally, b) the reverse edges that are added explicitly, and c) self loops. The S-GCN portion of  \reffig{fig:overview} represents this component.
%     $L(w_i, w_j) = \rightarrow$ if $(w_i, w_j, l(w_i, w_j)) \in E'$, i.e., if the edge is an original dependency parse edge
%    $L(w_i, w_j) = \leftarrow$ if $(w_i, w_j, l(w_i, w_j)^{-1}) \in E'$, i.e., if the edges is an inverse edge    
%    $L(w_i, w_j) = \top$ if $(w_i, w_j, \top) \in E'$, i.e., if the edge is a self-loop with $w_i = w_j$.

More formally, $H^{cntx} \in \mathbb{R}^{n \times k}$  is transformed to   $H^{syn} \in \mathbb{R}^{n \times k_{syn}}$  by applying S-GCN.
\subsubsection{Temporal Embedding}
\label{sec:time_embedding}
In this layer, NeuralDater exploits the Event-Time graph structure present in the document. %The Stanford’s SUTime tagger \cite{sutime} and CAEVO \cite{Chambers14} for date normalization and event extraction. 
CATENA \cite{catena_paper}, current state-of-the-art temporal and causal relation extraction algorithm, produces the temporal graph from the event time annotation of the document. GCN applied over this Event-Time graph, namely T-GCN, chooses $n_{T}$ number of tokens out of total $n$ tokens from the document for further revision in their embeddings. Note that T is the total number of events and time mentions present in the document. A special node DCT is added to the graph and its embedding is jointly learned. Note that this layer learns both label and direction specific parameters.

\subsubsection{Classifier}
Finally, the DCT embedding concatenated with the average pooled syntactic embedding is fed to a softmax layer for classification. This whole procedure is trained jointly.

%\item {\bf Incorporating Attention in Graph:}
%Include equations for incorporating graph attention from \cite{graphAttention} and also some ideas from \cite{graphsage}
%
%\item {\bf Attentive Neural Dater Overview}
	% !TeX spellcheck = en_GB
\section{Attentive Deep Document Dater (\method{}): Proposed Method}
\label{sec:model}

In this section, we describe Attentive Deep Document Dater (\method{}), our proposed method. \method{} is inspired by NeuralDater, and shares many of its components. Just like in NeuralDater, \method{} also leverages two main types of signals from the document -- syntactic and event-time --  to predict the document's timestamp. However, there are crucial differences between the two systems. Firstly, instead of concatenating embeddings learned from these two sources as in NeuralDater, \method{} treats these two models completely separate and combines them at a later stage. Secondly, unlike NeuralDater, \method{} employs attention mechanisms in each of these two models. We call the resulting models \methodsyntac{} (\methodsyntacshort) and \methodtemp{} (\methodtempshort{}). These two models are described in \refsec{sec:reader} and \refsec{sec:reasoner}, respectively.

%The main theme behind document dating can be broadly divided into two parts - context capturing and reasoning over temporal constraints. Although NeuralDater combines different imperative components in a principled way, it does not reflect on these two key ideas directly. In our work, we train two different models (\reffig{fig:overview}) separately and were able to show that they capture significantly different aspects of documents. Combination of these two models enables us to beat the state-of-the-art NeuralDater, with a substantially large margin of accuracy. We give a detailed description of the two components in the following subsections.

\subsection{\methodsyntac{} (\methodsyntacshort)}
\label{sec:reader}
%<<<<<<< HEAD

%A reader must be attentive enough to grasp the key-words from the content. 
Recent success of attention-based deep learning models for classification \cite{hierarchical_attn}, question answering \cite{attention_qna}, and  machine translation \cite{bahdanau_attn}  have motivated us to use attention during document dating. We extend the syntactic embedding model of NeuralDater (\refsec{sec: syntactic_embedding}) by incorporating an attentive pooling layer. %We try to mimic the behavior of reading through attentive pooling in the Syntactic-GCN (\refsec{sec: syntactic_embedding}). 
We call the resulting model \methodsyntacshort{}. This model (right side in \reffig{fig:overview}) has two major components.
\begin{itemize}
	\item\textbf{Context Embedding and Syntactic Embedding}:  Following NeuralDater, we used Bi-LSTM and S-GCN to capture context and long-range syntactic dependencies in the document (Please refer to \refsec{sec:context_lstm}, \refsec{sec: syntactic_embedding} for brief description). The syntactic embedding, $H^{syn} \in \mathbb{R}^{n \times k_{syn}}$ is then fed to an  Attention Network for further processing. Note that, $k_{syn}$ is the dimension of the output of Syntactic-GCN and $n$ is the number of tokens in the document.
	
	\item \textbf{Attentive Embedding}: In this layer, we learn the representation for the whole document through word level attention network. We learn a context vector, $u_{s} \in \mathbb{R}^{s}$ with respect to which we calculate attention for each token. Finally, we aggregate the token features with respect to their attention weights in order to represent the document. More formally, let $h^{syn}_{t} \in \mathbb{R}^{k_{syn}}$ be the  syntactic representation of the $t^{th}$ token in the document. We take non-linear projection of it in $\mathbb{R}^{s}$ with $W_{s} \in \mathbb{R}^{s \times k_{syn} }$. Attention weight $\alpha_{t}$ for  $t^{th}$ token is calculated with respect to the context vector $u_{t}^{T}$ as follows.
	$$u_{t} = \mathrm{tanh}(W_{s}h^{syn}_{t}),$$
	$$\alpha_{t} = \frac{\mathrm{exp}(u_{t}^{T}u_{s})}{\sum_{t}\mathrm{exp}(u_{t}^{T}u_{s})}.$$
	
	Finally, the document representation for the \methodsyntacshort{} is computed as shown below.
	\[
	  d_{\mathrm{\methodsyntacshort{}}} = \sum_{t}\alpha_{t}h^{syn}_{t}
	 \]
	  This representation is fed to a softmax layer for the final classification.
\end{itemize}
%<<<<<<< HEAD
The final probability distribution over years predicted by the \methodsyntacshort{} is given below.
$$\mathrm{P}_{\mathrm{\methodsyntacshort{}}}(y \vert D) = \mathrm{Softmax}(W \cdot d_{\mathrm{\methodsyntacshort{}}} + b).$$
%=======
%The final probability distribution over years predicted by the \methodsyntac{} for document D will be 
%$$\mathrm{P}_{\mathrm{\methodsyntacshort{}}}(y \vert D) = \mathrm{Softmax}(W \cdot d^{\mathrm{\methodsyntacshort{}}} + b).$$
%
%>>>>>>> 1510a819f7c0a1a42bd2d6d74ecfca7d587db998

\subsection{\methodtemp{} (\methodtempshort{})}
\label{sec:reasoner}

The \methodtempshort{} model is shown on the left side of \reffig{fig:overview}. Just like in \methodsyntacshort{}, context and syntactic embedding is also part of \methodtempshort{}. The syntactic embedding is fed to the Attentive Graph Convolution Network (AT-GCN) where the graph is obtained from the time-event ordering algorithm CATENA \cite{catena_paper}. We describe these components in detail below.

%The event extractor and automatic temporal graph creation algorithms, many times, output edges which are far from being perfect and there is no scope for us to validate that directly. The noise it produces contains a huge amount of potential risk for downstream predictions. Again, not every edge is important for the prediction, rather in most of the cases only a few temporal links are enough to constrain the DCT. Inspired  from the recent literature of improved feature aggregation of GCN  \cite{graphsage}, we use attentive graph convolution to handle the above mentioned hazards. Attentive graph convolution is a special way of doing convolution, where the contribution of the neighbouring nodes are weighted by attention weights, which are learnt.\\
%The context and syntactic embedding is also a part of the \methodtemp. The syntactic embedding is fed to the attentive graph convolution network where the  graph is obtained from the time-event ordering algorithm CATENA \cite{catena_paper}.

\subsubsection{Temporal Graph }

We use the same process used in NeuralDater \cite{NeuralDater} for procuring the Temporal Graph from the document. CATENA \cite{catena_paper} generates 9 different temporal links between events and time expressions present in the document. Following \citet{NeuralDater}, we choose 5 most frequent ones - AFTER, BEFORE, SIMULTANEOUS, INCLUDES, and IS INCLUDED -- as labels. The temporal graph is constructed from the partial ordering between event verbs and time expressions.

Let $\m{E}_{T}$ be the edge list of the Temporal Graph. Similar to \cite{gcn_srl,NeuralDater}, we also add reverse edges for each of the existing edge and self loops for passing current node information as explained in \refsec{sec:directed_gcn}. The new edge list $\m{E}_{T}^{'}$ is shown below.
\begin{multline*}    
\m{E'_{\text{T}}} = \m{E_{\text{T}}} \cup \{(j,i,l(i,j)^{-1})~|~ (i,j,l(i,j)) \in \m{E_{\text{T}}}\} \\
\cup \{(i, i, \mathrm{self})~|~i \in \m{V})\}.
%\label{eqn: updated_edges}
\end{multline*}
%

%<<<<<<< HEAD
%=======
 The reverse edges are added with reverse labels like $\mathrm{AFTER}^{-1}$, $\mathrm{BEFORE}^{-1}$ etc . Finally, we get 10 labels for our temporal graph and we denote the set of edge labels by $\m{L}$.
%>>>>>>> 1510a819f7c0a1a42bd2d6d74ecfca7d587db998
\subsubsection {Attentive Graph Convolution (AT-GCN)} 
\label{sec:graph_atten}

Since the temporal graph is automatically generated, it is likely to have incorrect edges. Ideally, we would like to minimize the influence of such noisy edges while computing temporal embedding. In order to suppress the noisy edges in the Temporal Graph and detect important edges for reasoning, we use attentive graph convolution \cite{graphsage} over the Event-Time graph. The attention mechanism learns the aggregation function jointly during training. Here, the main objective is to calculate the attention over the neighbouring nodes with respect to the current node for a given label. Then the embedding of the current node is updated by mixing neighbouring node embedding according to their attention scores. In this respect, we propose a label-specific attentive graph convolution over directed graphs.

Let us consider an edge in the temporal graph from node $i$ to node $j$ with type $l$, where $l \in \m{L}$ and $\m{L}$ is the label set. The label set $\m{L}$ can be divided broadly into two coarse labels as done in \refsec{sec: syntactic_embedding}. The attention weights are specific to only these two type of edges to reduce parameter and prevent overfitting. 
For illustration, if there exists an edge from node $i$  to $j$ then the edge types will be,
\begin{itemize}
	\item $ L(i, j) =\  \rightarrow$ if $(i, j, l(i, j)) \in \m{E}_{T}^{'}$,\\ i.e., if the edge is an original event-time edge.
	\item  $L(i, j) =\  \leftarrow$ if $(i, j, l(i, j)^{-1}) \in \m{E}_{T}^{'}$,\\ i.e., if the edge is added later.
\end{itemize}
First, we take a linear projection ($W^{atten}_{L(i,j)} \in \mathbb{R}^{F \times k_{syn} }$) of both the nodes in $\mathbb{R}^{F}$ in order to map both of them in the same direction-specific space. The concatenated vector $ [W^{atten}_{L(i,j)}\times h_{i}; W^{atten}_{L(i,j)} \times h_{j} ]$, signifies the importance of the node $j$ w.r.t. node $i$. A non linear transformation of this concatenation can be treated as the importance feature vector between $i$ and $j$.

$$ e_{ij} = \mathrm{tanh} [W^{atten}_{L(i,j)} \times h_{i}; W^{atten}_{L(i,j)} \times h_{j} ].$$
\begin{figure}[t]
	\centering
	\includegraphics[width=3in]{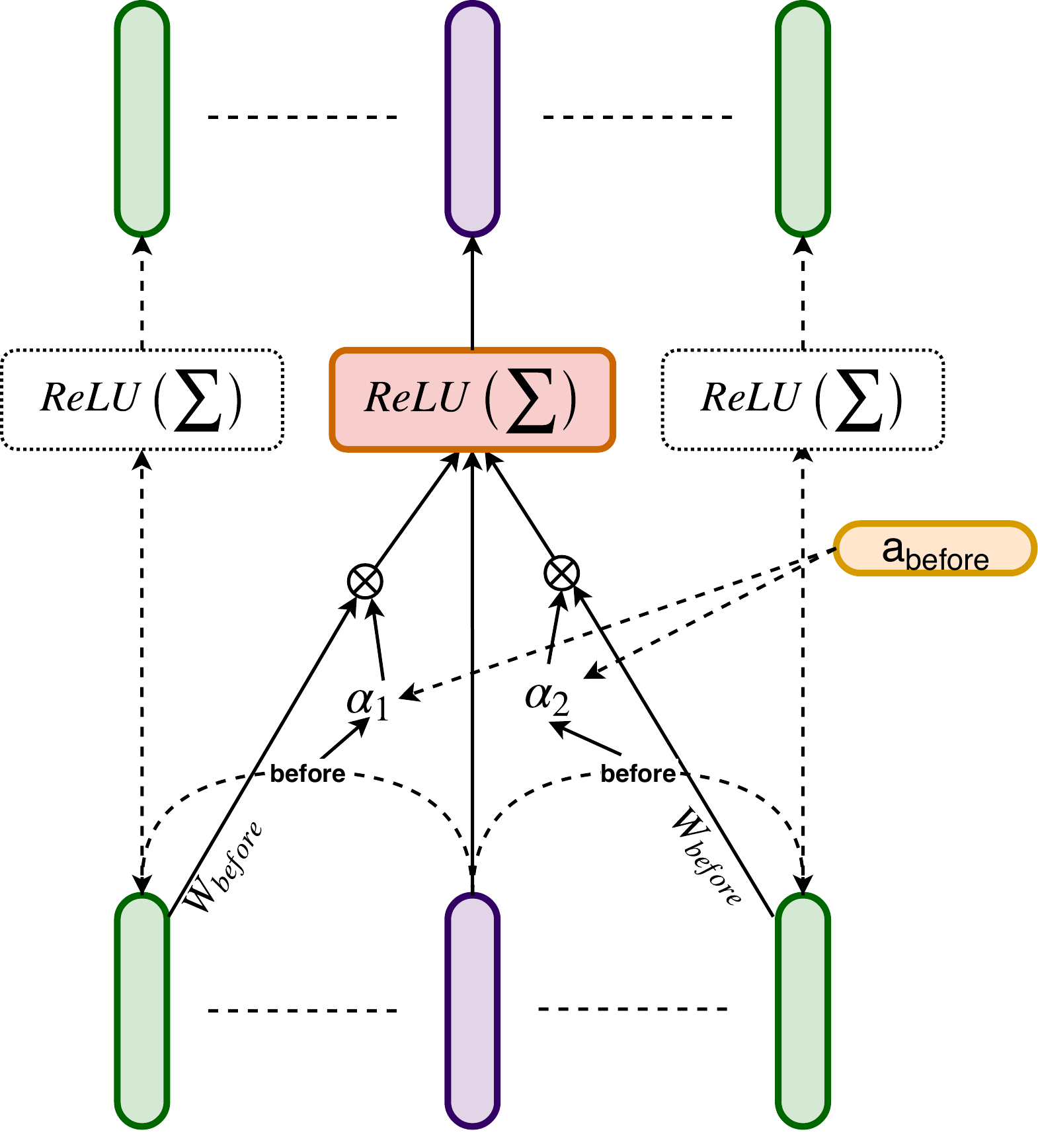}
	\caption{\label{fig:Attentive_Graph_Convolution} Attentive Graph Convolution (AT-GCN). In this layer, we learn attention weights for every edge based on label and direction. The attention weights are learnt using a context vector. The final representation of every node is a summation of weighted convolution over neighboring nodes based on labels.%Evaluating performance of different methods on dating  documents with and without time mentions.
	}
\end{figure}
%<<<<<<< HEAD
%Now, we compute the attention weight of node $j$ for node $i$  with respect to a context vector \reminder{a is L based}$a \in \mathbb{R}^{2F}$ as follows.  
%\[   
%	\alpha^{l(i, j)}_{ij} = \frac{\text{exp}\left(a^{T} e_{ij}   \right) }{\sum\limits_{  k \in \m{N}^{l(i,\cdot)}_{i}}\text{exp} \left(a^{T} e_{ik}   \right)  },
%=======
Now, we compute the attention weight of node $j$ for node $i$  with respect to a direction-specific context vector $a_{L(i,j)} \in \mathbb{R}^{2F}$, as follows.  
\[
	\alpha^{l(i, j)}_{ij} = \frac{\text{exp}\left(a_{L(i,j)}^{T} e_{ij}   \right) }{\sum\limits_{  k \in \m{N}^{l(i,\cdot)}_{i}}\text{exp} \left(a_{L(i,j)}^{T} e_{ik}   \right)  },
%>>>>>>> 1510a819f7c0a1a42bd2d6d74ecfca7d587db998
\label{eqn:attention_cal}
\]
where, $\alpha^{l(i, j)}_{ij} = 0$ if node $i$ and $j$ is not connected through label $l$. $\m{N}^{l(i,\cdot)}$ denotes the subset of the neighbourhood of node $i$ with label $l$ only. Please note that, although the linear transform weight ($ W^{atten}_{L(i,j)}  \in \mathbb{R}^{F \times k_{syn}}$) is specific to the coarse labels $L$, but for each finer label $l \in \m{L}$ we get these convex weights of attentions. \reffig{fig:Attentive_Graph_Convolution} illustrates the above description w.r.t. edge type BEFORE.

Finally, the feature aggregation is done according to the attention weights. Prior to that, another label specific linear transformation is taken to perform the convolution operation. Then, the updated feature for node $i$ is calculated %carried out 
as follows.
\[
\Scale[0.88]{ h_{i}^{k+1} = f \left( \sum_{l \in \m{L}} \sum_{j \in \m{N}^{l(i,\cdot)}_{i}}\alpha^{l(i, j)}_{ij}\left(W_{l(i,j)}h_{j} + b_{l(i,j)}\right)\right).
\label{eqn:graph_atten}}
\]
where, $ \alpha_{ii} = 1$, $\ \m{N}^{l(i,\cdot)}$ denotes the subset of the neighbourhood of node $i$ with label $l$ only. Note that,  $\alpha^{l(i, j)}_{ij} = 0$ when $j \notin \m{N}^{l(i,\cdot)}$. To illustrate formally, from \reffig{fig:Attentive_Graph_Convolution}, we see that weight $\alpha_{1}$ and $\alpha_{2}$ is calculated specific to label type BEFORE and the neighbours which are connected through BEFORE is being multiplied with $W_{before}$ prior to aggregation in the $ReLU$ block.

Now, after applying attentive graph convolution network, we only consider the representation of Document Creation Time (DCT),  $h_{DCT}$, as the document representation itself. $h_{DCT}$ is now passed through a fully connected layer prior to softmax. Prediction of the \methodtempshort{} for the document D will be given as
$$\mathrm{P}_{\mathrm{\methodtempshort}}(y \vert D) = \mathrm{Softmax}(W \cdot d_{\mathrm{DCT}} + b).$$

\subsection{\method{}: Attentive Deep Document Dater} 
\label{sec:read_reason}

\begin{figure}[t]
	\centering
	\includegraphics[width=3in]{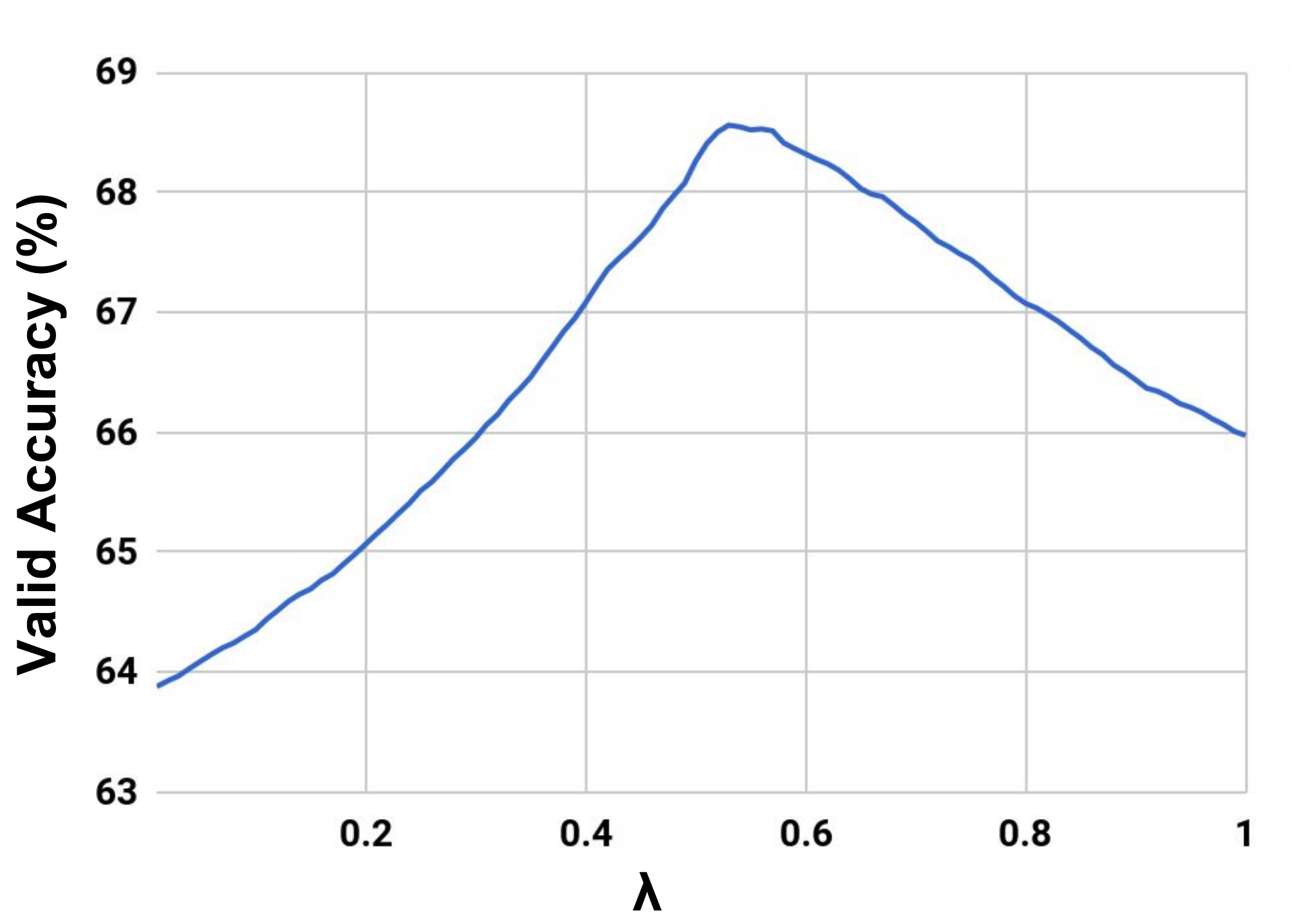}
	\caption{\label{fig:Accuracy_vs_Lambda} Variation of validation accuracy with $\lambda$ (for APW dataset). We observe that \methodsyntacshort{} and \methodtempshort{} are both important for the task as we get optimal $\lambda$ = 0.52.
	}
\end{figure}

In this section, we propose an unified model by mixing both \methodsyntacshort{} and \methodtempshort{}. Even on validation data, we see that performance of both %(\ref{sec:reasoner} and \ref{sec:reader})
 the models differ to a large extent. This significant difference (McNemar test p $ < $ 0.000001) motivated the unification. We take convex combination of the output probabilities of the two models as shown below.
\begin{multline*}
$$\mathrm{P}_{joint}(y \vert D) = \lambda\mathrm{P}_{\mathrm{\methodsyntacshort}}(y \vert D) \\
+(1-\lambda) \mathrm{P}_{\mathrm{\methodtempshort}}(y \vert D)$$.
\end{multline*}

The combination hyper-parameter $\lambda$ is tuned on the validation data. %For different datasets, 
We obtain the value of $\lambda$ to be 0.52 (\reffig{fig:Accuracy_vs_Lambda}) and 0.54 for APW and NYT datasets, respectively. This depicts that the two models are capturing significantly different aspects of documents, resulting in a substantial improvement in performance when combined.

	% !TeX spellcheck = en_US
\section{Experimental Setup}
\label{sec:experiments}

\begin{table}[t]
	\begin{tabular}{cccc}
		\toprule
		Datasets 	& \# Docs & Start Year & End Year\\
		\midrule
		APW 		&  675k	& 1995  & 2010 \\
		NYT			&  647k	& 1987  & 1996 \\
		\bottomrule
		\addlinespace
	\end{tabular}
	\caption{\label{tb:datasets}Details of datasets used. Please refer \refsec{sec:experiments} for details.}
\end{table}

%\subsection{Datasets}
%\label{sec:datasets}
\textbf {Dataset}: Experiments are carried out on the Associated Press Worldstream (APW) and New York Times (NYT) sections of the Gigaword corpus \cite{gigaword5th}. We have used the same 8:1:1 split as \citet{NeuralDater} for all the models. %, which is available in the supplementary section. 
For quantitative details please refer to Table \ref{tb:datasets}.

\textbf{Evaluation Criteria}: In accordance with prior work \cite{Chambers:2012:LDT:2390524.2390539, Kotsakos:2014:BAD:2600428.2609495,NeuralDater} the final task is to predict the publication year of the document. %, \cite{}. 
We give a brief description of the baselines below. 

\textbf{Baseline Methods:}
\begin{itemize}
	\item \textbf{MaxEnt-Joint} \cite{Chambers:2012:LDT:2390524.2390539}: This method engineers several hand-crafted temporally influenced features to classify the document using MaxEnt Classifier.
	\item \textbf{BurstySimDater} \cite{Kotsakos:2014:BAD:2600428.2609495}:  
	 This is a purely statistical method which uses lexical similarity and term burstiness \cite{Lappas:2009:BSD:1557019.1557075} for dating documents in arbitrary length time frame. For our experiments, we used a time frame length of 1 year.
	\item \textbf{NeuralDater} \cite{NeuralDater}: This is the first deep neural network based approach for the document dating task. Details are provided in \refsec{sec:neural_dater}.	
\end{itemize}

\textbf{Hyperparameters:} We use 300-dimensional GloVe embeddings and 128-dimensional hidden state for both GCNs and BiLSTM with 0.8 dropout. We use Adam \cite{adam_optimizer} with 0.001 learning rate for training. For \methodtempshort{} we use  2-layers of AT-GCN. 1-layer of S-GCN is used for both the models.
%\textbf {Our Proposed Methods:}
%\begin{itemize}
%	\item\textbf{ \methodsyntac{}}, which exploits the contextual information in documents using attention over weights. (refer  \refsec{sec:reader})
%	\item \textbf{\methodtemp{}}, which exploits the temporal graph structure of document using attentive graph convolution. (refer \refsec{sec:reasoner})
%	\item \textbf{Joint model}, which ensembles the previous two models in a principled fashion and outperforms all the previous baseline methods. (refer \refsec{sec:read_reason}) 
%\end{itemize}
\begin{figure}[t]
	\centering
	\includegraphics[width=3.1in]{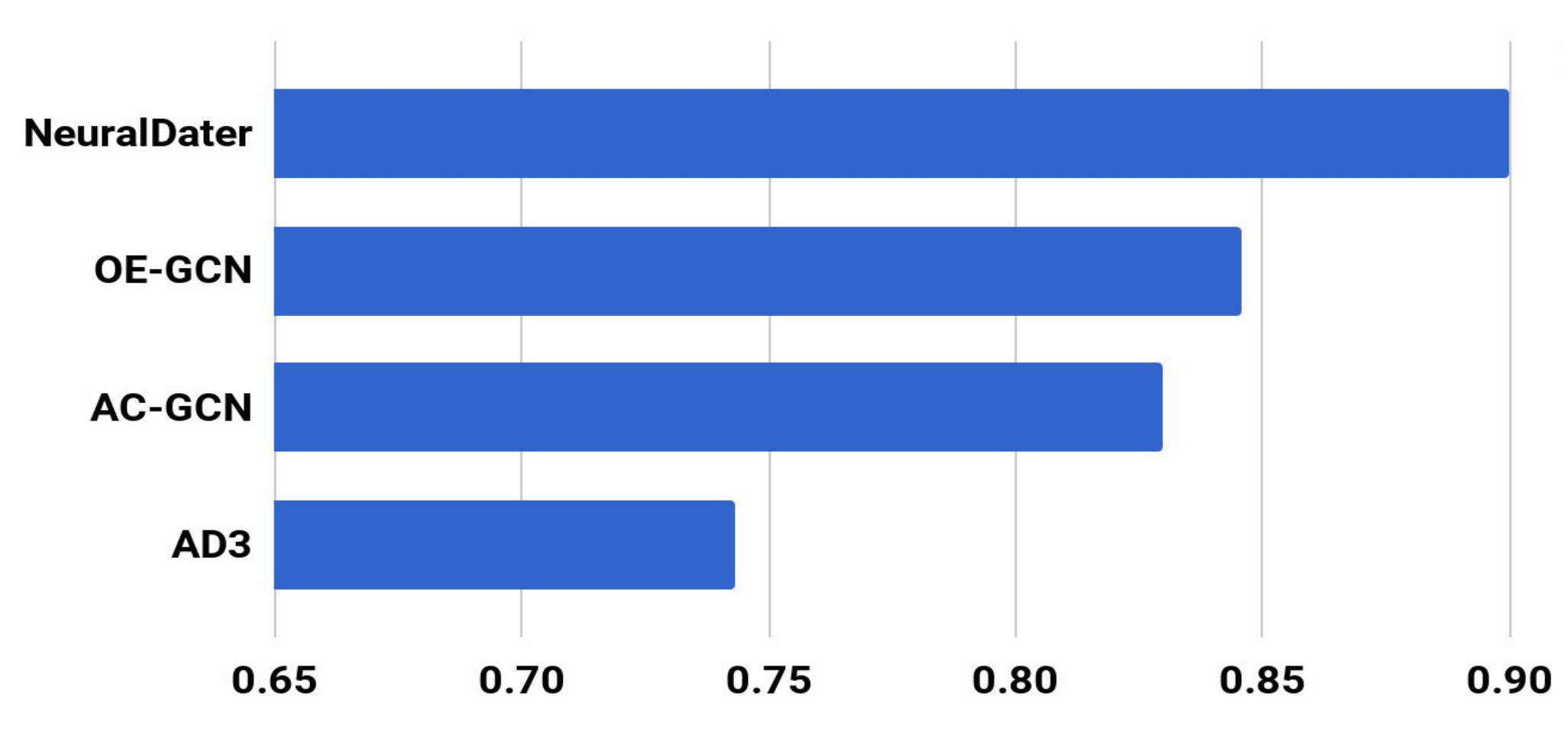}
	\caption{\label{fig:results_mean_dev}Mean absolute deviation (in years; lower is better) between a model's top prediction and the true year in the APW dataset. We find that all of our proposed methods outperform the previous state-of-the-art NeuralDater. Please refer to \refsec{sec:perf} for details. %Evaluating performance of different methods on dating  documents with and without time mentions.
	}
\end{figure}

%\begin{figure}[t]
%	\centering
%	\includegraphics[width=3.1in]{images/Time_Mention.pdf}
%	\caption{\label{fig:Accuracy_vs_time_mention}Accuracy of various models for with and without time-mentions in document. Our proposed models outperform the previous baseline models. %Evaluating performance of different methods on dating  documents with and without time mentions.
%	}
%\end{figure}
	% !TeX spellcheck = en_US

\begin{table}[t]
	\centering
	\begin{tabular}{lcc}
		\toprule
		Method 			 & APW & NYT \\
		\midrule		
		\addlinespace
		BurstySimDater 		& 45.9 & 38.5 \\
		MaxEnt-Joint		& 52.5 & 42.5 \\
		NeuralDater			& 64.1 & 58.9 \\
		\hline
		\addlinespace
		\AND{} [\ref{sec:atten_perf}] & 66.2 & 60.1\\
		\methodtempshort{} [\ref{sec:reasoner}]& 63.9 & 58.3\\
		\methodsyntacshort{}  [\ref{sec:reader}]& 65.6 & 60.3\\
%		Joint & 66.1&60.5\\
		\hline
		\addlinespace
		\textbf{\method{}} [\ref{sec:read_reason}] & \textbf{68.2} & \textbf{62.2} \\
		\bottomrule
	\end{tabular}
	\caption{\label{tb:result_main}Accuracy (\%) of different methods on the APW and NYT datasets for the document dating problem (higher is better). The unified model significantly outperforms all previous models.}
\end{table}

\begin{table}[t]
	\centering
	\begin{tabular}{lcc}
		\toprule
		Method 			 & Accuracy \\
		\midrule		
		\addlinespace
		T-GCN of NeuralDater			& 61.8  \\
		\methodtempshort				 & \textbf{63.9} \\
		\midrule		
		\addlinespace
		S-GCN of NeuralDater          &63.2\\
		\methodsyntacshort 				& \textbf{65.6}\\
		\bottomrule
	\end{tabular}
	\caption{\label{tb:atten_result}Accuracy (\%) comparisons of component models with and without Attention. This results show the effectiveness of both word attention and Graph Attention for this task. Please see \refsec{sec:atten_perf} for more details.}
\end{table}
%\begin{figure}[t]
%	\centering
%	\includegraphics[width=3in]{images/mean_abs_dev.pdf}
%	\caption{\label{fig:results_mean_dev}Mean absolute deviation (in years; lower is better) between a model's top prediction and the true year in the APW dataset. We find that Neural Dater, the proposed method, achieves the least deviation. Please see \ref{sec:perf_comp} for details. %Evaluating performance of different methods on dating  documents with and without time mentions.
%	}
%\end{figure}

%\begin{table}[!t]
%	\begin{small}
%		\centering
%		\begin{tabular}{lc}
%			\toprule
%			Method 			 & Accuracy \\
%			\midrule		
%			\addlinespace
%			%		CNN									& 56.3 \\
%			T-GCN 								& 57.3 \\
%			S-GCN + T-GCN $(K=1)$				& 57.8 \\
%			S-GCN + T-GCN $(K=2)$				& 58.8 \\
%			S-GCN + T-GCN $(K=3)$				& \textbf{59.1} \\
%			\midrule
%			Bi-LSTM 							& 58.6 \\
%			Bi-LSTM + CNN 						& 59.0 \\
%			Bi-LSTM + T-GCN						& 60.5 \\
%			Bi-LSTM + S-GCN + T-GCN (no gate)	& 62.7 \\
%			Bi-LSTM + S-GCN + T-GCN $(K=1)$		& \textbf{64.1} \\
%			Bi-LSTM + S-GCN + T-GCN $(K=2)$		& 63.8 \\
%			Bi-LSTM + S-GCN + T-GCN $(K=3)$		& 63.3 \\
%			\bottomrule
%		\end{tabular}
%	\caption{\label{tb:result_ablation}Accuracies  of different ablated methods on the APW dataset. Overall, we observe that incorporation of context (Bi-LSTM), syntactic structure (S-GCN) and temporal structure (T-GCN) in Neural Dater achieves the best performance. Please see \ref{sec:perf_comp} for details.}
%\end{small}
%\end{table}

\section{Results}
\label{sec:results}
\subsection{Performance Analysis}
\label{sec:perf}
\begin{figure}[t!]
	\centering
	\includegraphics[width=3in]{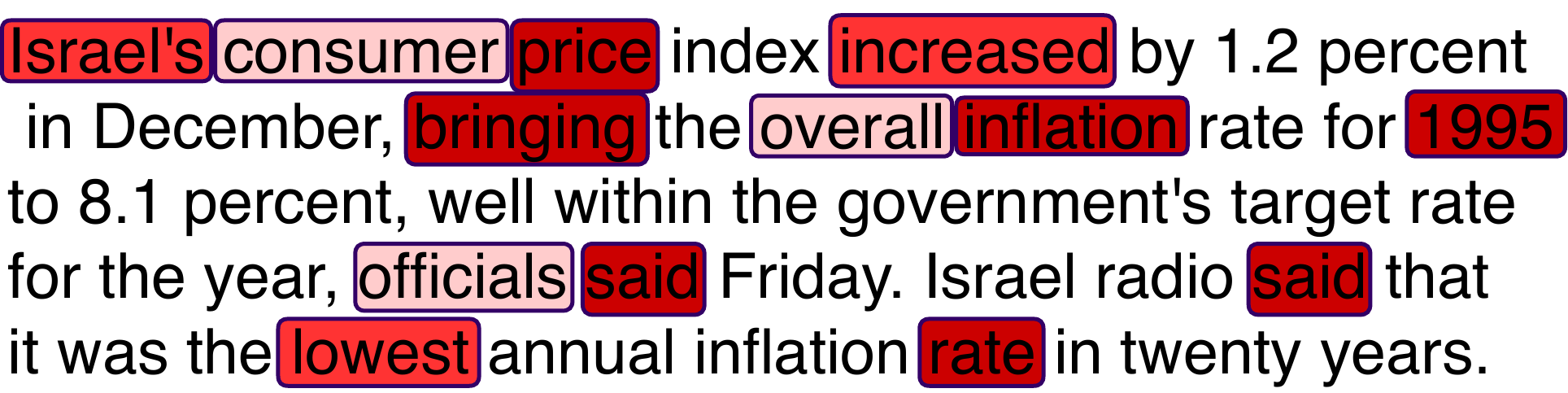}
	\caption{\label{fig:Context_Attention} Visualization of the attention of \methodsyntacshort{}. \methodsyntacshort{} captures the intuitive tokens as seen in the figure. Darker shade implies higher attention. The correct DCT is 1996.%Evaluating performance of different methods on dating  documents with and without time mentions.
	}
\end{figure}
\begin{figure}[t!]
	\centering
	\includegraphics[width=3in]{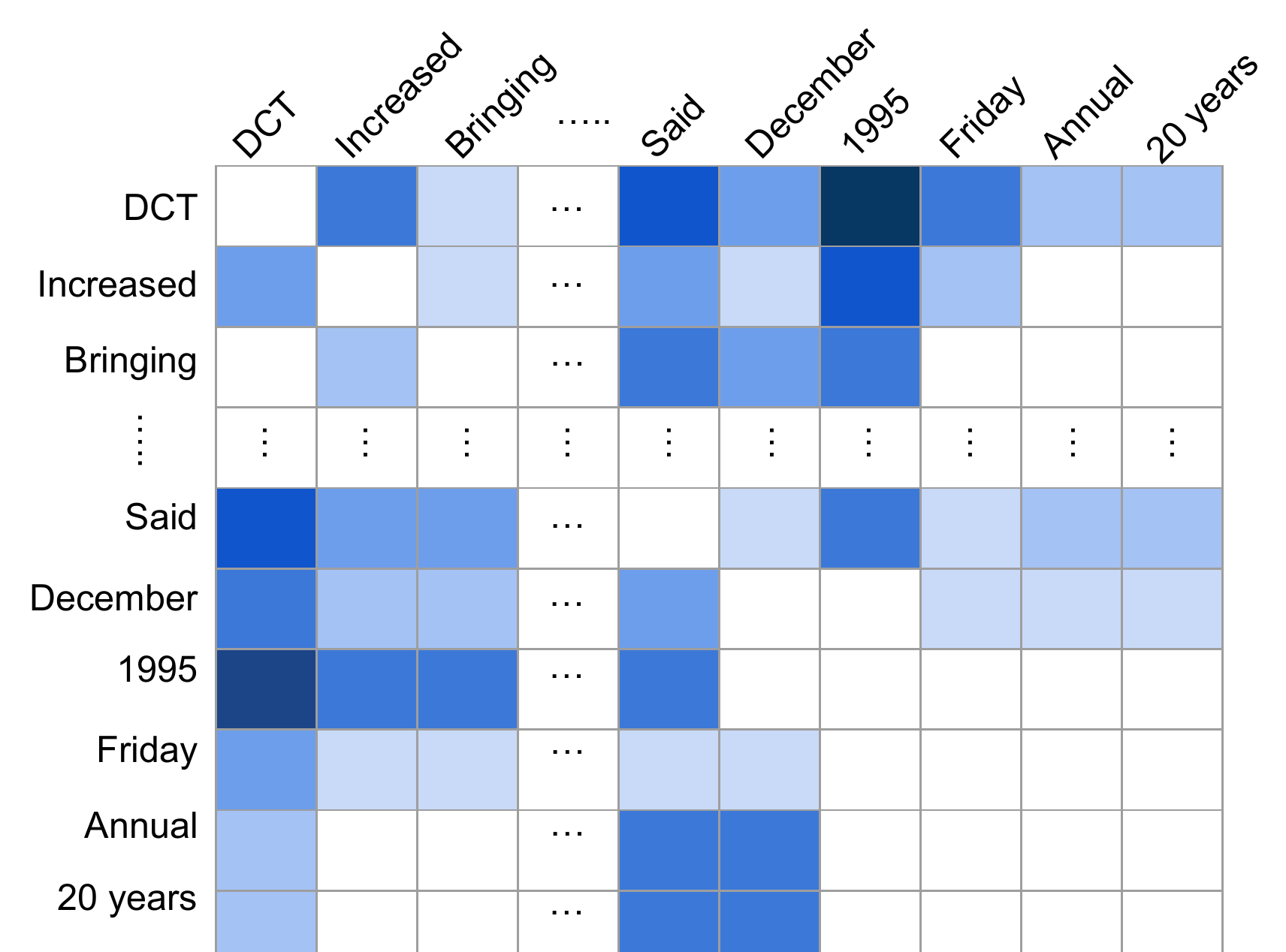}
	\caption{\label{fig:DCT_Attention} Visualization of the average edge attention of the temporal graph as learnt by \methodtempshort{} for the document shown in \reffig{fig:Context_Attention}. Darker color implies higher attention. The correct DCT is 1996.%Evaluating performance of different methods on dating  documents with and without time mentions.
	}
\end{figure}
In this section, we compare the effectiveness of our method with that of prior work. The deep network based NeuralDater model in \cite{NeuralDater}  outperforms previous feature engineered \cite{Chambers:2012:LDT:2390524.2390539} and statistical methods \cite{Kotsakos:2014:BAD:2600428.2609495} by a large margin. We observe a similar trend in our case. Compared to the state-of-the-art model NeuralDater, we gain, on an average, a 3.7\% boost in accuracy on both the datasets (\reftbl{tb:result_main}).

Among individual models, \methodtempshort{} performs at par with NeuralDater, while \methodsyntacshort{} outperforms it. The empirical results imply that \methodsyntacshort{} by itself is effective for this task. The relatively worse performance of \methodtempshort{} can be attributed to the fact that it only focuses on the Event-Time information and leaves out most of the contextual information. However, it captures various different (p $<$ 0.000001, McNemar’s test, 2-tailed) aspects of the document for classification, which motivated us to propose an ensemble of the two models. This explains the significant boost in performance of \method{} over NeuralDater as well as the individual models. It is worth mentioning that although \methodsyntacshort{} and \methodtempshort{} do not provide significant boosts in accuracy, their predictions have considerably lower mean-absolute-deviation as shown in \reffig{fig:results_mean_dev}.

We concatenated the DCT embedding provided by \methodtempshort{} with the document embedding provided by \methodsyntacshort{} and trained in an end to end joint fashion like NeuralDater. We see that even with a similar training method, the \AND{} model on an average, performs 1.6\% better in terms of accuracy, once again proving the efficacy of attention based models over normal models.

\subsection{Effectiveness of Attention}
\label{sec:atten_perf}
Attentive Graph Convolution (\refsec{sec:graph_atten}) proves to be effective for \methodtempshort{}, giving a 2\% accuracy improvement over non-attentive T-GCN of NeuralDater (Table \ref{tb:atten_result}). Similarly the efficacy of word level attention is also prominent from \reftbl{tb:atten_result}.\\
We have also analyzed our models by visualizing attentions over words and attention over graph nodes. Figure \ref{fig:Context_Attention} shows that %\methodsyntac{} 
AC-GCN focuses on temporally informative words such as "said" (for tense) or time mentions like ``1995'', alongside important contextual words like ``inflation'', ``Israel'' etc. For \methodtempshort{}, from \reffig{fig:DCT_Attention} we observe that ``DCT'' and time-mention `1995' grabs the highest attention. Attention between ``DCT'' and other event verbs indicating past tense are quite prominent, which helps the model to infer 1996 (which is correct) as the most likely time-stamp of the document. These analyses provide us with a good justification for the performance of our attentive models.
%\subsection{Qualitative Results}
%\label{sec:quality}
%\subsubsection{Visualization of attention} In order show that out, we visualize the attention layers for Graph as well as words. \reminder{include examples and then describe them briefly}

%\subsubsection{Effectiveness of \methodtemp{} with Increasing Time expression}

	% !TeX spellcheck = en_US
\section{Discussion}
\label{sec:discussion}

\begin{figure}[t!]
	\centering
	\includegraphics[width=2.25in]{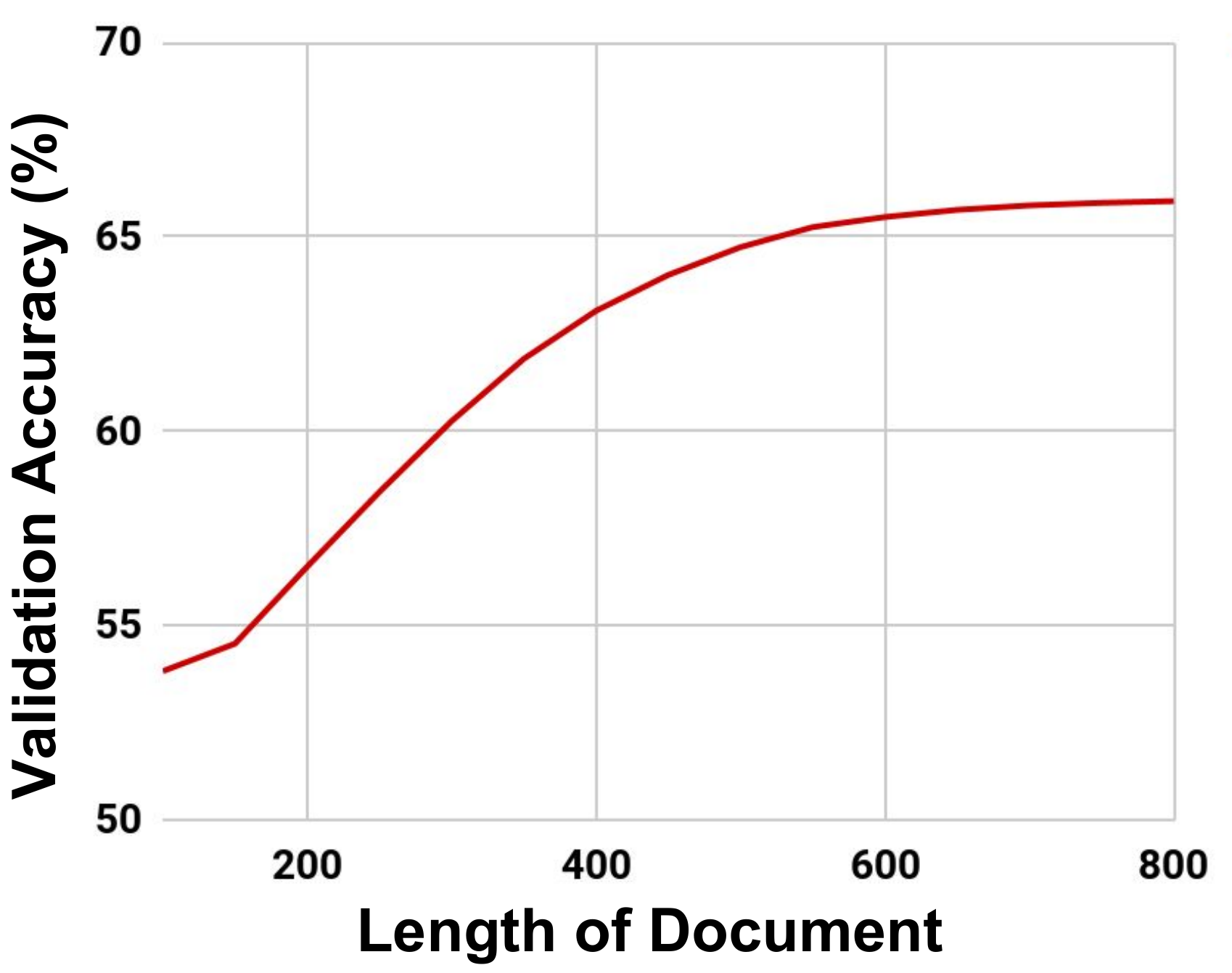}
	\caption{\label{fig:context_vs_length} Variation of validation accuracy (\%) with respect to length of documents (for APW dataset) for \methodsyntacshort{}. Documents having more than 100 tokens are selected for this analysis. Please see \refsec{sec:discussion}.
	}
\end{figure}

\begin{figure}[t!]
	\centering
	\includegraphics[width=2.25in]{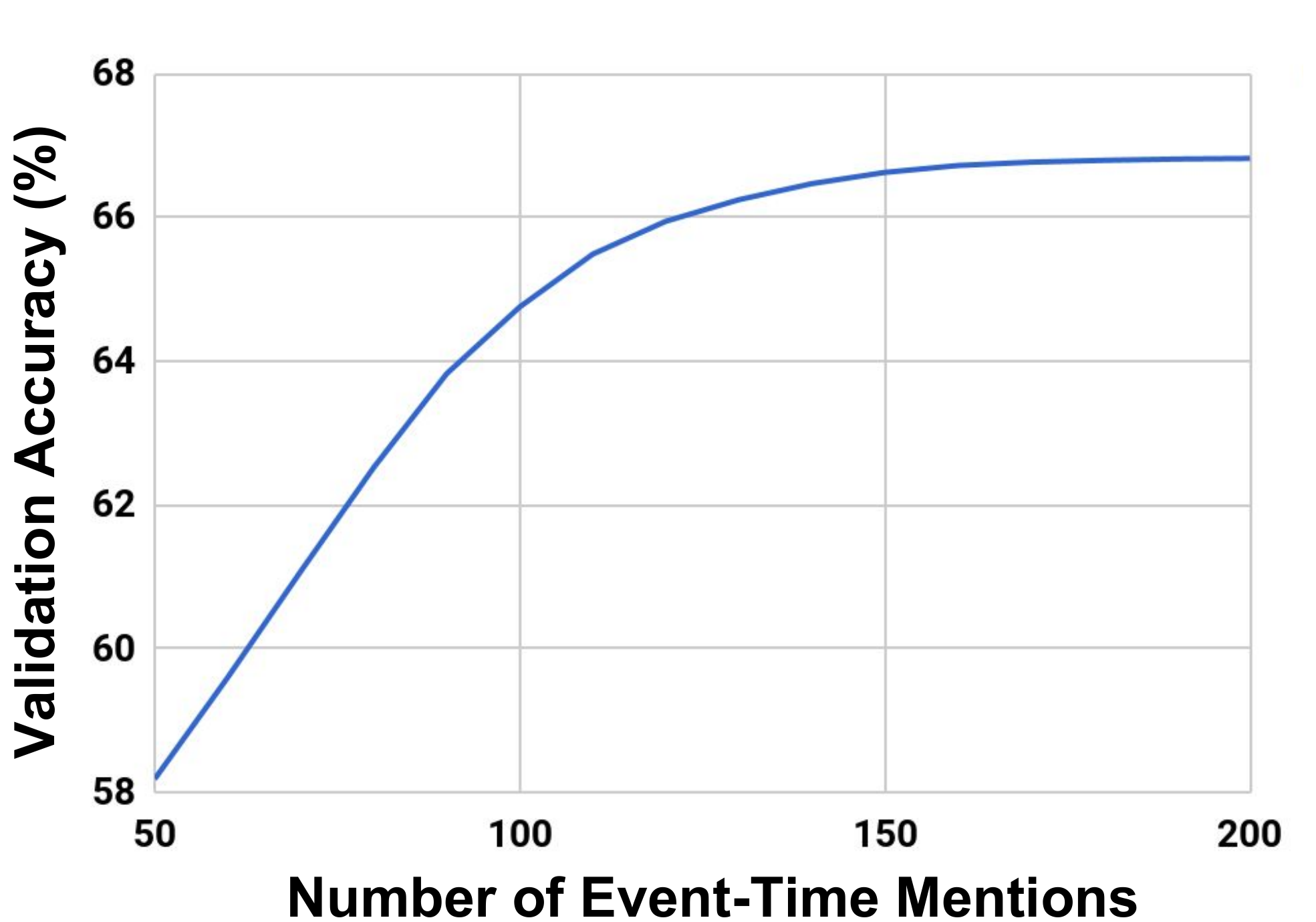}
	\caption{\label{fig:DCTpref_with_ET} Variation of validation accuracy (\%) with number of event-time mentions in documents (for APW dataset) for \methodtempshort{}. Documents with more than 100 tokens are selected for this analysis. Please see \refsec{sec:discussion}. %Evaluating performance of different methods on dating  documents with and without time mentions.
	}
\end{figure}

Apart from empirical improvements over previous models, we also perform a qualitative analysis of the individual models. Figure \ref{fig:context_vs_length} shows that the performance of \methodsyntacshort{} improves with the length of documents, thus indicating that richer context leads to better model prediction.  \reffig{fig:DCTpref_with_ET} shows how the performance of \methodtempshort{} improves with the number of event-time mentions in the document, thus further reinforcing our claim that more temporal information improves model performance.

\citet{NeuralDater} reported that their model got confused by the presence of multiple misleading time mentions. \method{} overcomes this limitation using attentive graph convolution, which successfully filters out noisy time mentions as is evident from  \reffig{fig:DCTpref_with_ET}. %We have successfully tied up these loose ends using attentive graph convolution, which successfully filters out noisy time mentions as is evident from  \reffig{fig:DCTpref_with_ET}. 

\section{Conclusion}
\label{sec:conclusion}

We propose \method{}, an ensemble model which exploits both syntactic and temporal information in a document explicitly to predict its creation time (DCT). To the best of our knowledge, this is the first application of attention based deep models for dating documents. Our experimental results demonstrate the effectiveness of our model over all previous models. We also visualize the attention weights to show that the model is able to choose what is important for the task and filter out noise inherent in language. %We believe further research is required to successfully solve this problem and towards that goal 
As part of future work, we would like to incorporate external knowledge as a side information for improved time-stamping of documents.

\section*{Acknowledgments}
\label{sec:acknowledgments}

%We thank the anonymous reviewers for their constructive comments. 
This work is supported by the Ministry of Human Resource Development (MHRD), Government of India.
	\balance
	%% The file named.bst is a bibliography style file for BibTeX 0.99c
%	\pagebreak
	\bibliography{references}
	\bibliographystyle{acl_natbib}
\end{document}